\documentclass[lettersize,journal]{IEEEtran}
\usepackage{amsmath,amsfonts}
\usepackage{algorithmic}
\usepackage{algorithm}
\usepackage{array}
\usepackage{textcomp}
\usepackage{stfloats}
\usepackage{url}
\usepackage{verbatim}
\usepackage{graphicx}
\usepackage{booktabs} 
\usepackage{cite}
\usepackage{multirow}
\usepackage{makecell}
\usepackage{bbding}
\usepackage{circledsteps} 
\usepackage{subcaption}
\usepackage{wrapfig}
\usepackage{xcolor}
\usepackage[table]{xcolor}
\usepackage{amssymb}

\usepackage[colorlinks,linkcolor=red,anchorcolor=black,citecolor=green]{hyperref}
\usepackage{orcidlink}

\begin{document}

\title{Long-Tail Adaptive Flow Matching with Explicit Conditional Consistency Guidance for Precise Multimodal Face Synthesis}

\author{Yushe Cao,  Xuechao Zou, Yuhui Chen, Xing Xi, Dianxi Shi, \\ Chun Yu\orcidlink{0000-0003-2591-7993},~\IEEEmembership{Senior Member, IEEE}, Junliang Xing\orcidlink{0000-0001-6801-0510	
},~\IEEEmembership{Senior Member, IEEE}
\thanks{Yushe Cao, Chun Yun, and Junliang Xing are with the Department of Computer Science and Technology, Tsinghua University, Beijing 100084, China (Email: cao-ys23@mails.tsinghua.edu.cn, chunyu@tsinghua.edu.cn, jlxing@tsinghua.edu.cn).}
\thanks{Yuhui Chen is with the Ant Group, Hangzhou, 310000, China (Email: chenyuhui.cyh@antgroup.com)}
\thanks{Xuechao Zou is with the School of Computer Science and Technology, Beijing Jiaotong University, Beijing 100044, China (Email: xuechaozou@foxmail.com)}
\thanks{Xing Xi is with the School of Computer Science and Engineering, South China University of Technology, Guangzhou 510006, China (Email: 202311089252@mail.scut.edu.cn).}
\thanks{Dianxi Shi is with the Intelligent Game and Decision Lab,  Beijing 100071, China(Email: dxshi@nudt.edu.cn).}
\thanks{Corresponding authors: Junliang Xing, Dianxi Shi.}}

\markboth{Preprint.~Under review}
{Shell \MakeLowercase{\textit{et al.}}: A Sample Article Using IEEEtran.cls for IEEE Journals}


\maketitle

\begin{abstract}
Although diffusion-based methods have substantially improved the controllability of multimodal face synthesis, their semantic alignment remains suboptimal because most existing approaches rely on implicit latent-space objectives to model the relationship between denoising variables and multimodal conditions. Such implicit modeling is often insufficient to enforce precise correspondence between synthesized faces and conditional inputs, especially under long-tailed semantic mask distributions where rare attributes receive weak optimization signals. To address these limitations, we propose EC\textsuperscript{2}Face, a multimodal face synthesis framework that improves semantic alignment through explicit semantic supervision and distribution-aware optimization. First, we introduce Explicit Conditional Consistency Guidance (ECCG), which imposes direct consistency supervision in pixel space by decoding an approximate reverse estimate of the clean latent and explicitly aligning the synthesized image with textual descriptions and semantic masks. A temporal dynamic modulation function is further designed to adapt the supervision strength according to the timestep-dependent reliability of reverse estimation. Second, we propose Long-Tail Adaptive Flow Matching (LAFM), which reweights spatial optimization signals based on semantic attribute frequency, with normalized weights to maintain numerical stability during training. Importantly, all additional modules are used only during training and introduce no extra inference overhead. Extensive experiments show that EC\textsuperscript{2}Face consistently outperforms competitive baselines in both generation quality and semantic alignment, achieving a 29.38\% improvement in mask accuracy on rare attributes.
\end{abstract}

\section{Introduction}
\label{sec:intro}
\IEEEPARstart{M}{ultimodal} face synthesis \cite{ning2023multi,melnik2024face,sowmya2024generative,meng2025mm2latent} aims to generate photorealistic and semantically controllable facial images under the joint guidance of heterogeneous conditions, such as text descriptions and semantic masks. Compared with unimodal face generation \cite{wu2023high,diao2025ft2tf,wang2025facea}, multimodal synthesis offers finer-grained control over facial attributes, local structures, and global appearance, and is therefore highly relevant to a broad range of applications, including portrait editing, digital human creation, and interactive visual content generation.

Early multimodal face synthesis methods were primarily built upon generative adversarial networks (GANs) \cite{liu2023gan,du2023pixelface+,meng2025mm2latent}. Although these methods demonstrated encouraging controllability under specific settings, they often suffered from training instability, limited diversity, and mode collapse, which restricted their ability to simultaneously achieve high fidelity and precise semantic control. More recently, diffusion models have emerged as the dominant paradigm for controllable image generation, owing to their progressive denoising mechanism and flexible conditioning interface \cite{mou2024t2i,peng2024controlnext,po2024state,he2025diffusion,chang2025design,cao2026multivariate}. In particular, Diffusion Transformers (DiTs) model multimodal inputs as unified tokens and enable global cross-modal interaction through self-attention, providing a stronger backbone for multimodal conditional synthesis \cite{peebles2023scalable,tan2025ominicontrol}.

\begin{figure} 
    \centering
    \includegraphics[width=1.0\linewidth]{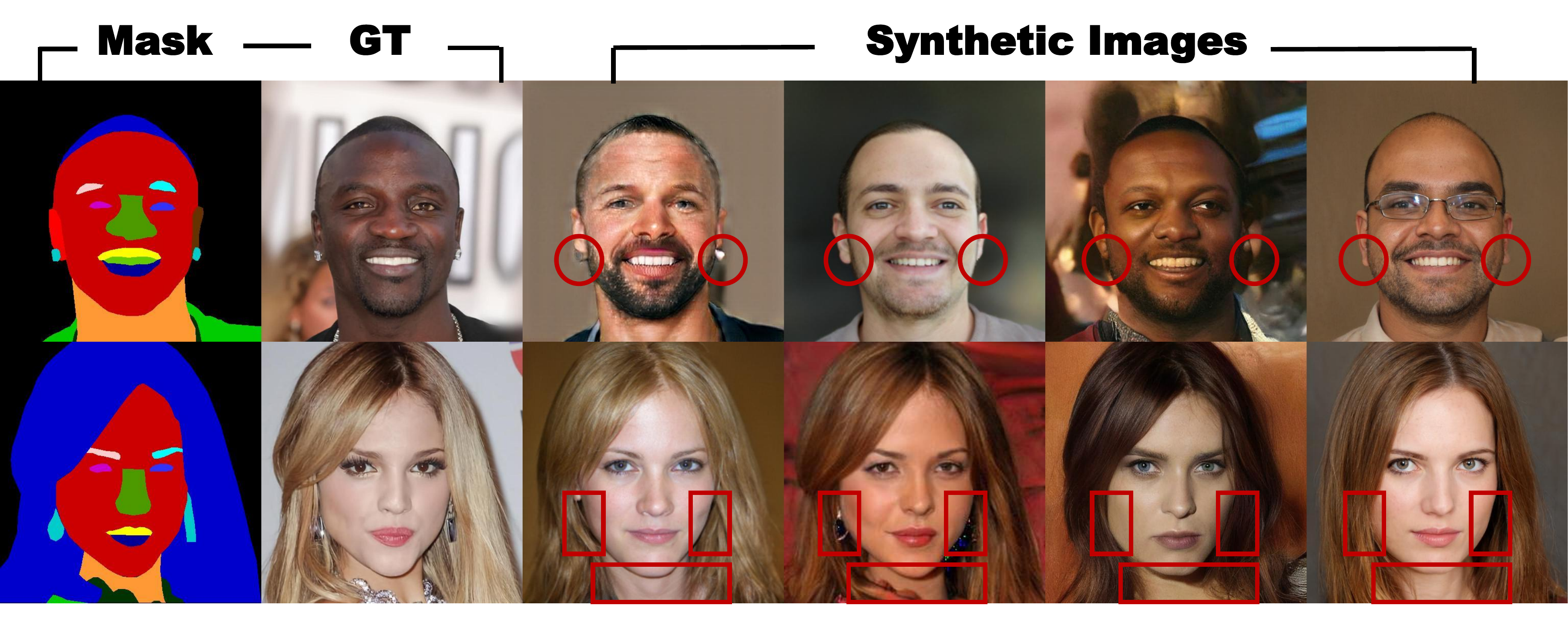}
    \caption{Common failure cases of existing multimodal face synthesis methods under long-tailed semantic masks. Rare attributes such as earrings and necklaces are often under-modeled or entirely omitted, even when they are explicitly specified in the mask condition.}
    \label{fig:common_issue}
\end{figure}

Despite these advances, existing multimodal diffusion models still face two critical bottlenecks when precise semantic alignment is required. First, most methods rely on denoising or flow matching objectives in latent space \cite{rombach2022high,dao2023flow}, where the relationship between noisy latents and multimodal conditions is learned only implicitly. While such objectives are effective for fitting data dynamics, they do not directly enforce the final generated image to be semantically consistent with the input text or structurally aligned with the input mask. As a result, generation may still exhibit semantic drift, local omission, or insufficient response to conditional cues. Second, semantic attributes in face parsing masks typically follow a pronounced long-tailed distribution. Frequent categories such as skin, hair, and background dominate optimization, whereas rare and spatially small attributes, such as earrings and necklaces, contribute only weak supervision signals. This imbalance biases the model toward high-frequency regions and substantially weakens its ability to faithfully synthesize rare attributes as exemplified by the cases shown in Figure \ref{fig:common_issue}.

To overcome these limitations, we propose EC\textsuperscript{2}Face, a multimodal face synthesis framework tailored for precise semantic alignment. Our key idea is to improve multimodal generation from two complementary perspectives: explicit conditional supervision and distribution-aware optimization. First, we introduce {Explicit Conditional Consistency Guidance} (ECCG), which imposes direct semantic supervision in pixel space. Specifically, an approximate clean latent is obtained via one-step reverse estimation and decoded into an image approximation, which is then explicitly aligned with the input text and semantic mask using pretrained semantic models. Since this reverse estimation is approximate and becomes less reliable at large timesteps, we further introduce a temporal dynamic modulation function to adapt the supervision strength according to timestep reliability, thereby improving alignment while maintaining training stability. Second, we propose {Long-Tail Adaptive Flow Matching} (LAFM) to mitigate optimization bias caused by long-tailed semantic attribute distributions. LAFM assigns frequency-aware spatial weights to the flow matching objective, allowing rare semantic regions to receive stronger optimization signals. To preserve numerical stability, the reweighting matrix is normalized before being applied to the loss, ensuring that the overall loss magnitude remains comparable to that of the original flow matching objective. Notably, both ECCG and LAFM are applied only during training and introduce no additional inference overhead, making EC\textsuperscript{2}Face a practical training-time enhancement for diffusion- or flow-based generation backbones.

The main contributions of this work are summarized as follows:
\begin{itemize}
    \item We propose EC\textsuperscript{2}Face, a multimodal face synthesis framework for precise semantic alignment, which improves controllable generation by combining explicit semantic supervision with distribution-aware optimization.

    \item We introduce ECCG, which enforces pixel-space semantic agreement between synthesized images and multimodal conditions. A temporal dynamic modulation function is further designed to adapt the supervision strength according to the timestep-dependent reliability of reverse estimation.

    \item We develop LAFM, which mitigates long-tail optimization bias through frequency-aware spatial reweighting with normalized weights, improving rare-attribute synthesis while preserving training stability.

    \item Extensive experiments demonstrate that EC\textsuperscript{2}Face outperforms strong baselines in generation quality, text-image consistency, mask alignment, and rare-attribute synthesis. All additional components are used only during training, incurring no extra inference cost.
\end{itemize}

\section{Related Work}
\subsection{Face Generation}

Face generation has been extensively studied under different conditional settings \cite{e3facenet,styleclip,chen2024dreamidentity,meng2025mm2latent}. Early research mainly focused on unimodal face generation, where a single modality such as text or semantic masks is used as the control signal. Text-driven methods~\cite{clip2latent,gcdp,e3facenet,song2025attridiffuser} enable semantic manipulation of facial attributes through natural language descriptions, offering flexible control over appearance-related factors such as hairstyle, expression, and accessories. In parallel, mask-guided approaches~\cite{inade,e2style,semflow,cao2024mefusion} emphasize explicit spatial controllability by specifying facial structures and region layouts with semantic masks. Although both paradigms have achieved promising results, each modality alone has intrinsic limitations: text-only generation often lacks precise spatial grounding, whereas mask-only generation is less effective in flexibly controlling fine-grained appearance attributes beyond structural layouts.

Generative Adversarial Networks (GANs), represented by the StyleGAN series~\cite{stylegan,stylegan2}, have played an important role in early controllable face synthesis, and StyleCLIP~\cite{styleclip} further extended GAN-based generation to language-guided semantic editing. To overcome the limitations of unimodal control, recent studies have increasingly explored multimodal face generation. Methods such as TediGAN~\cite{tedigan} and PixelFace+~\cite{pixelfaceplus} combine textual and visual conditions, including masks or sketches, to achieve more comprehensive control over identity, pose, and facial attributes. Hybrid frameworks integrating GANs with diffusion-based generators~\cite{ddgi,uniteandconquer} further improve cross-modal interaction and generation diversity. Nevertheless, existing multimodal face generation methods still struggle with precise semantic alignment across heterogeneous conditions, especially for attributes associated with small or infrequent regions. In practice, dominant facial regions often contribute disproportionately to optimization, while long-tail semantic attributes such as earrings remain insufficiently modeled. This limitation motivates us to explicitly enhance the learning of under-represented semantic regions during training.

\begin{figure*}[h]
    \centering
    \includegraphics[width=\linewidth]{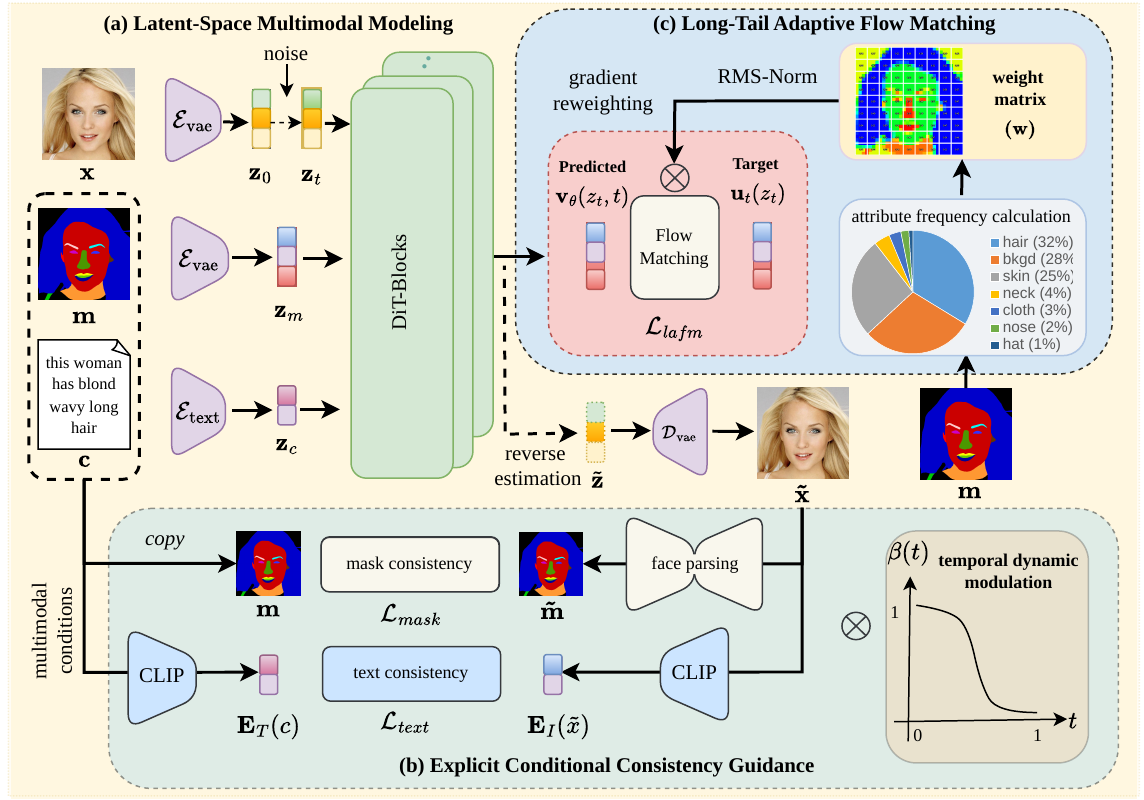}
    \caption{
       Overview of EC\textsuperscript{2}Face: A training framework devised to enhance precise semantic alignment in multimodal face synthesis. Building upon Latent-Space Multimodal Modeling, this framework incorporates two pivotal improvements—Explicit Conditional Consistency Guidance and Long-Tail Adaptive Flow Matching—which markedly boost the alignment accuracy between synthesized facial images and the input textual descriptions as well as semantic masks. 
    }
    \label{fig:framework}
\end{figure*}

\subsection{Conditional Diffusion and Flow Matching}

The paradigm of generative modeling has gradually shifted from GAN-based methods to diffusion models~\cite{ddpm,ddim,ldm,cao2026dual}, which show clear advantages in training stability, generation quality, and sample diversity. DDPM~\cite{ddpm} established the standard denoising diffusion framework, while DDIM~\cite{ddim} improved sampling efficiency through deterministic non-Markovian trajectories. Latent Diffusion Models (LDM)~\cite{ldm} further reduced computational overhead by performing diffusion in latent space, making high-resolution conditional generation more practical. Building on these advances, Classifier-Free Guidance (CFG)~\cite{cfg} enables an effective trade-off between fidelity and conditional consistency, and ControlNet as well as its variants~\cite{controlnet,peng2024controlnext} significantly strengthen controllability by injecting structural conditions into pretrained diffusion models.

Recent studies have also introduced transformer-based generative architectures~\cite{peebles2023scalable,wang2025lavin,tan2025ominicontrol,gligen}, which improve the modeling of long-range dependencies and multimodal interactions, as well as flow-based formulations such as flow matching~\cite{dao2023flow}, which directly learn continuous transport dynamics between source and target distributions. Despite these advances, whether based on diffusion denoising~\cite{ddpm} or flow matching~\cite{dao2023flow}, most existing methods still correlate latent variables with multimodal conditions in an implicit and globally averaged manner. Such optimization is often dominated by easy regions and easy samples, limiting fine-grained consistency between generated results and multimodal conditions under challenging cases. Consequently, semantically critical but difficult conditions may not receive sufficient optimization emphasis during training. This observation motivates us to improve multimodal face generation from two complementary perspectives: enhancing the optimization of under-represented semantic regions with LAFM, and strengthening hard-case multimodal consistency with ECCG.


\section{Method}
\label{sec:method}

EC\textsuperscript{2}Face is built upon a DiT backbone for multimodal face synthesis with precise semantic alignment. As illustrated in Fig.~\ref{fig:framework}, it consists of three key components: Latent-Space Multimodal Modeling, Explicit Conditional Consistency Guidance, and Long-Tail Adaptive Flow Matching. The first serves as the base generation backbone, while ECCG and LAFM are two complementary training-time enhancements for improving semantic alignment and rare-attribute modeling, respectively. Importantly, the two newly introduced modules are utilized solely during the training phase and do not incur any additional inference overhead.

\subsection{Latent-Space Multimodal Modeling}
\label{sec:latent_modeling}

\textbf{Notation.}
Let $x \in \mathbb{R}^{H \times W \times 3}$ denote a face image, $m \in \{1,\ldots,n\}^{H \times W}$ its semantic mask with $n$ predefined categories, and $c$ the text condition. A Variational Autoencoder (VAE) encoder maps $x$ into the latent space, yielding the clean latent $z_0$. Given a timestep $t \in [0,1]$, the corresponding noisy latent is denoted by $z_t$. To model conditional generation, we adopt a DiT to predict the conditional velocity field
$v_{\theta}(z_t, m, c, t),$
which is conditioned on the noisy latent $z_t$, semantic mask $m$, text input $c$, and timestep $t$. This latent-space formulation provides the basis for multimodal face synthesis.

\textbf{Multimodal Modeling.}
To enable unified cross-modal interaction, the text condition, mask condition, and noisy image latent are embedded into a shared token space. Denoting the corresponding token sequences by $z_c$, $z_m$, and $z_t$, the multi-head attention operation is defined as
\begin{equation}
\mathrm{Attn}(q,k,v)
=
\mathrm{softmax}\left(\frac{qk^\top}{\sqrt{d_h}}\right)v,
\end{equation}
where $d_h$ is the dimension of each attention head. The query, key, and value matrices are constructed via modality-specific linear projections:
\begin{equation}
\begin{aligned}
q &= [W_q^{(c)}z_c;\, W_q^{(t)}z_t;\, W_q^{(m)}z_m],\\
k &= [W_k^{(c)}z_c;\, W_k^{(t)}z_t;\, W_k^{(m)}z_m],\\
v &= [W_v^{(c)}z_c;\, W_v^{(t)}z_t;\, W_v^{(m)}z_m],
\end{aligned}
\label{eq:tri_attn}
\end{equation}
where $W_{*}^{(c)}$, $W_{*}^{(t)}$, and $W_{*}^{(m)}$ denote learnable projection matrices for text, image, and mask tokens, respectively.

Although this latent-space modeling enables effective multimodal generation, latent-space flow matching alone does not explicitly enforce semantic consistency between synthesized images and input conditions. In addition, under long-tailed semantic mask distributions, standard optimization tends to favor frequent regions and under-optimize rare attributes. To address these issues, EC\textsuperscript{2}Face introduces ECCG and LAFM, which are described below.

\subsection{Explicit Conditional Consistency Guidance}
\label{sec:eccg}

Most existing DiT-based generation models \cite{flux2024,tan2025ominicontrol} optimize only latent-space objectives, which provide conditional supervision only implicitly through denoising dynamics. However, such implicit learning is often insufficient for precise semantic alignment, especially when the generated image needs to simultaneously satisfy textual descriptions and semantic mask constraints. To alleviate this issue, we propose ECCG, which introduces direct semantic supervision in pixel space.

\textbf{One-Step Reverse Estimation.}
Given a noisy latent $z_t$ and the model-predicted velocity field $v_{\theta}(z_t, m, c, t)$, we first estimate an approximation of the clean latent by a one-step reverse update:
\begin{equation}
    \tilde{z} = z_t - t \cdot v_{\theta}(z_t, m, c, t).
    \label{eq:reverse_estimate}
\end{equation}
The approximated latent $\tilde{z}$ is then decoded by the VAE decoder to obtain an image approximation $\tilde{x}$ in pixel space. This approximation serves as an intermediate image representation on which explicit multimodal consistency constraints can be imposed.

\textbf{Text Consistency Loss.}
To measure semantic consistency between the generated image approximation $\tilde{x}$ and the text condition $c$, we employ a pretrained CLIP model~\cite{radford2021learning}, including an image encoder $\mathbf{E}_I$ and a text encoder $\mathbf{E}_T$. We define the text consistency loss using cosine distance:
\begin{equation}
\mathcal{L}_{\text{text}}
=
1-\frac{\mathbf{E}_I(\tilde{x}) \cdot \mathbf{E}_T(c)}
{\|\mathbf{E}_I(\tilde{x})\|\,\|\mathbf{E}_T(c)\|}.
\label{eq:text_loss}
\end{equation}

\textbf{Mask Consistency Loss.}
To enforce structural consistency with the semantic mask condition, we feed $\tilde{x}$ into a pretrained face parsing model FaRL~\cite{zheng2022general} and obtain its predicted semantic map $\tilde{m}$. We then compute the mask consistency loss as a weighted combination of cross-entropy loss and Dice loss~\cite{milletari2016v}:
\begin{equation}
\begin{aligned}
\mathcal{L}_{\text{mask}}
&=
\lambda \, \mathcal{L}_{\text{CE}}(\tilde{m},m)
+
(1-\lambda)\,\mathcal{L}_{\text{Dice}}(\tilde{m},m) \\
&=
\lambda \left[
-\sum_i \left(
m_i \log \tilde{m}_i + (1-m_i)\log(1-\tilde{m}_i)
\right)\right] \\
&\quad +
(1-\lambda)\left(
1-\frac{2\sum_i m_i\tilde{m}_i}
{\sum_i m_i^2 + \sum_i \tilde{m}_i^2 + \delta}
\right),
\end{aligned}
\label{eq:mask_loss}
\end{equation}
where $\lambda \in [0,1]$ balances the two terms and $\delta > 0$ is a smoothing constant for numerical stability. In our experiments, we set $\lambda=0.5$.

\begin{figure}[t]
    \centering
    \includegraphics[width=\linewidth]{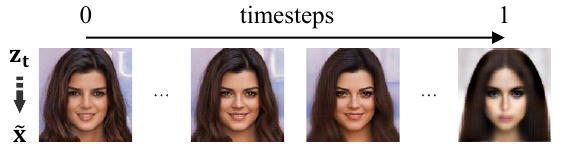}
    \caption{Reliability of one-step reverse estimation decreases as the timestep $t$ increases.}
    \label{fig:reverse_estimate}
\end{figure}

\textbf{Temporal Dynamic Modulation.}
The effectiveness of ECCG depends on the quality of the image approximation $\tilde{x}$, which is derived from the one-step reverse estimate in Eq.~\eqref{eq:reverse_estimate}. However, this approximation is not equally reliable across timesteps. As shown in Fig.~\ref{fig:reverse_estimate}, its reliability degrades as $t$ increases. Directly applying strong pixel-space supervision at large timesteps may therefore introduce noisy gradients and impair training stability.

To address this issue, we design a temporal dynamic modulation function to adjust the supervision strength according to timestep reliability:
\begin{equation}
    \beta(t)=\frac{1}{1+\exp\left(\gamma_s(t-t_c)\right)},
    \label{eq:t_modu_func}
\end{equation}
where $\gamma_s$ controls the slope and $t_c$ determines the inflection point. This design assigns larger weights to low-timestep samples, where reverse estimation is more reliable, and suppresses supervision at high timesteps. In practice, we set $\gamma_s=10$ and $t_c=0.4$.

\begin{figure*}[h]
    \centering
        \includegraphics[width=0.98\linewidth]{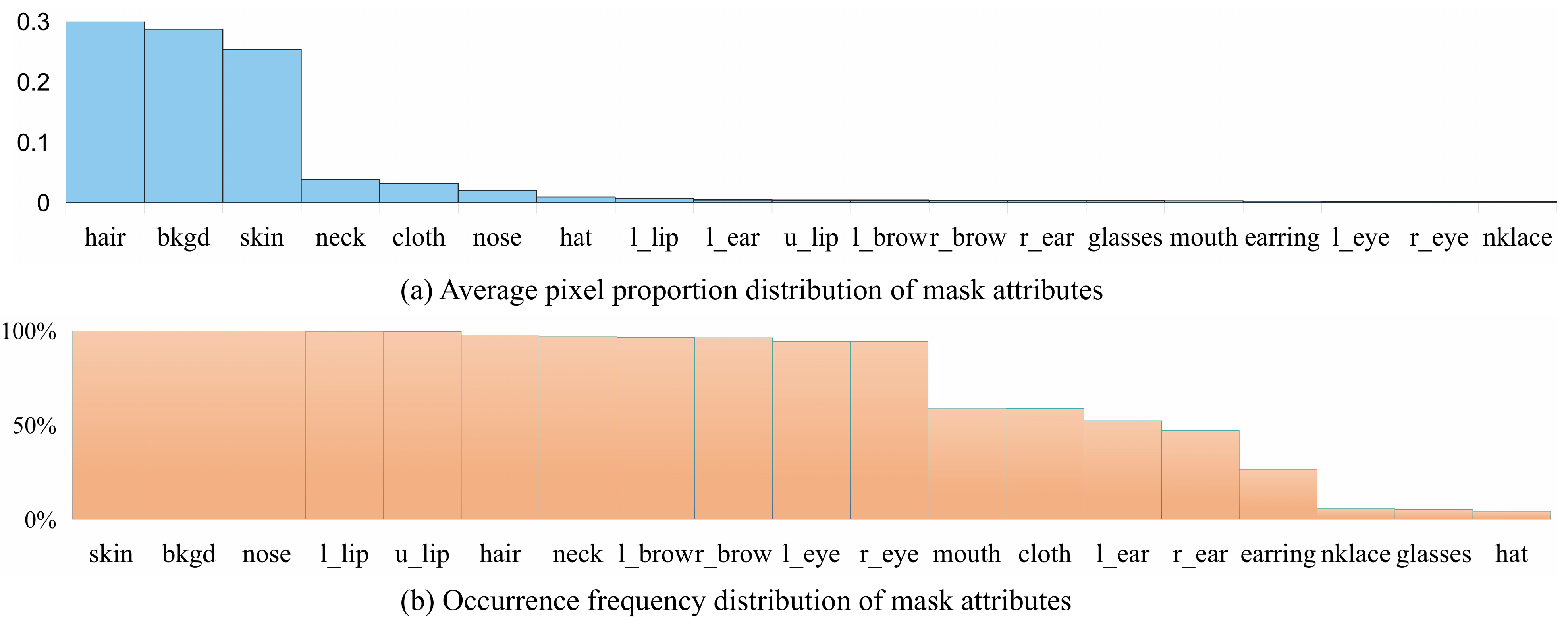}
    \caption{Long-tail distribution phenomenon of predefined mask attributes in the MM-CelebA-HQ dataset.}
    \label{fig:attribute_distribution}
\end{figure*}

\textbf{Multimodal Consistency Loss.}
Combining the text and mask consistency terms, the final ECCG loss is defined as
\begin{equation}
    \mathcal{L}_{\text{eccg}}
    =
    \beta(t)\left(
    \alpha_1 \mathcal{L}_{\text{text}}
    +
    \alpha_2 \mathcal{L}_{\text{mask}}
    \right),
    \label{eq:muliti_condition_consist_loss}
\end{equation}
where $\alpha_1$ and $\alpha_2$ are balancing coefficients. In this way, ECCG explicitly encourages semantic agreement between generated images and multimodal conditions in pixel space, complementing the implicit supervision provided by latent-space flow matching.

\subsection{Long-Tail Adaptive Flow Matching}
\label{sec:lafm}

A practical challenge in multimodal face synthesis is the severe imbalance of semantic attributes in facial masks. As illustrated by our statistical analysis of the commonly used MM-CelebA-HQ~\cite{lee2020maskgan} dataset in Figure \ref{fig:attribute_distribution}, dominant regions such as skin, hair, and background occupy most pixels and appear in almost every sample, whereas attributes such as earrings and necklaces are both spatially sparse and infrequent. Under such a long-tailed distribution, standard flow matching tends to be dominated by frequent semantic regions, resulting in insufficient optimization for rare attributes. To alleviate this issue, we propose LAFM, which introduces distribution-aware spatial reweighting into the latent-space flow matching objective. The core idea is to amplify training signals in rare semantic regions while suppressing the overwhelming influence of high-frequency categories.

\textbf{Sample-Wise Distribution-Aware Spatial Weighting.}
In practice, rather than estimating global statistics over the entire dataset, we compute semantic statistics for each training sample based on its mask $m$. Specifically, let $N_k(m)$ denote the number of pixels belonging to category $k$ in the current sample. We then define a sample-wise spatial weight matrix $w \in \mathbb{R}^{H \times W}$ as
\begin{equation}
w(i,j)=\frac{\sum_{k=1}^{n} N_k(m)}{n \cdot N_{m(i,j)}(m)+1},
\label{eq:weight_matrix}
\end{equation}
where $m(i,j)$ denotes the semantic label at position $(i,j)$. In this way, the proposed distribution-aware reweighting is instantiated through sample-wise frequency-aware weighting: semantic categories occupying fewer pixels in the current sample receive larger weights, while dominant regions are relatively downweighted.

\textbf{RMS-Normalized Reweighting.}
To keep the weighted objective on a scale comparable to the original flow matching loss, we normalize the weight matrix by its root mean square (RMS):
\begin{equation}
\bar{w}(i,j)
=
\frac{w(i,j)}
{\sqrt{\frac{1}{HW}\sum_{u=1}^{H}\sum_{v=1}^{W} w(u,v)^2}},
\label{eq:rms_weight}
\end{equation}
where $H$ and $W$ denote the spatial height and width, respectively. This normalization satisfies
\begin{equation}
\frac{1}{HW}\sum_{i=1}^{H}\sum_{j=1}^{W}\bar{w}(i,j)^2=1,
\end{equation}
which stabilizes optimization while preserving the relative emphasis on under-represented semantic regions.

\textbf{Reweighted Flow Matching Objective.}
Let $\mu_t(z_t)$ denote the target velocity field at timestep $t$ under the adopted flow matching path. The final LAFM objective is defined as
\begin{equation}
\mathcal{L}_{\text{lafm}}
=
\mathbb{E}_{t,\epsilon,z}
\left[
\frac{1}{HW}
\left\|
\bar{w}\odot
\left(
v_\theta(z_t,m,c,t)-\mu_t(z_t)
\right)
\right\|_2^2
\right],
\label{eq:opt-flm}
\end{equation}
where $\odot$ denotes element-wise multiplication. By explicitly increasing the contribution of spatially sparse semantic regions during training, LAFM alleviates optimization bias toward dominant regions and improves the modeling of long-tail facial attributes, without introducing any additional inference-time overhead.

\subsection{Training Strategy}
\label{sec:training}

\textbf{LoRA Fine-tuning.}
To reduce computational cost and parameter redundancy, we adopt Low-Rank Adaptation (LoRA)~\cite{hu_lora:_2021} for parameter-efficient fine-tuning. Specifically, LoRA modules are inserted into the visual embedding layer and selected linear layers of the transformer blocks, while the remaining pretrained parameters are frozen. This design introduces less than 0.1\% additional trainable parameters compared with full fine-tuning. The overall training objective consists of the ECCG loss and the LAFM loss:
\begin{equation}
\begin{aligned}
    \mathcal{L}_{\text{total}}
    &= \mathcal{L}_{\text{lafm}} + \mathcal{L}_{\text{eccg}} \\
    &= \mathcal{L}_{\text{lafm}} + \beta(t)\left(\alpha_1 \mathcal{L}_{\text{text}} + \alpha_2 \mathcal{L}_{\text{mask}}\right).
\end{aligned}
\label{eq:total_loss}
\end{equation}

\textbf{Stochastic Condition Dropout.}
To enable both multimodal collaborative generation and unimodal generation within a unified model, we randomly drop the text condition $c$ and the mask condition $m$ independently with probability $p$ during training ($p=0.1$ in our experiments):
\begin{equation}
\begin{aligned}
    c/m =
    \begin{cases}
        \phi, & \text{with probability } p,\\
        c/m, & \text{with probability } 1-p.
    \end{cases}
\end{aligned}
\label{eq:random_drop}
\end{equation}
Here, $\phi$ denotes an empty condition. This strategy improves robustness to missing conditions and enables flexible text-only, mask-only, and joint conditional generation.

\section{Experiments}
\label{sec:expriments}
\begin{figure*}[t]
    \centering    \includegraphics[width=\linewidth]{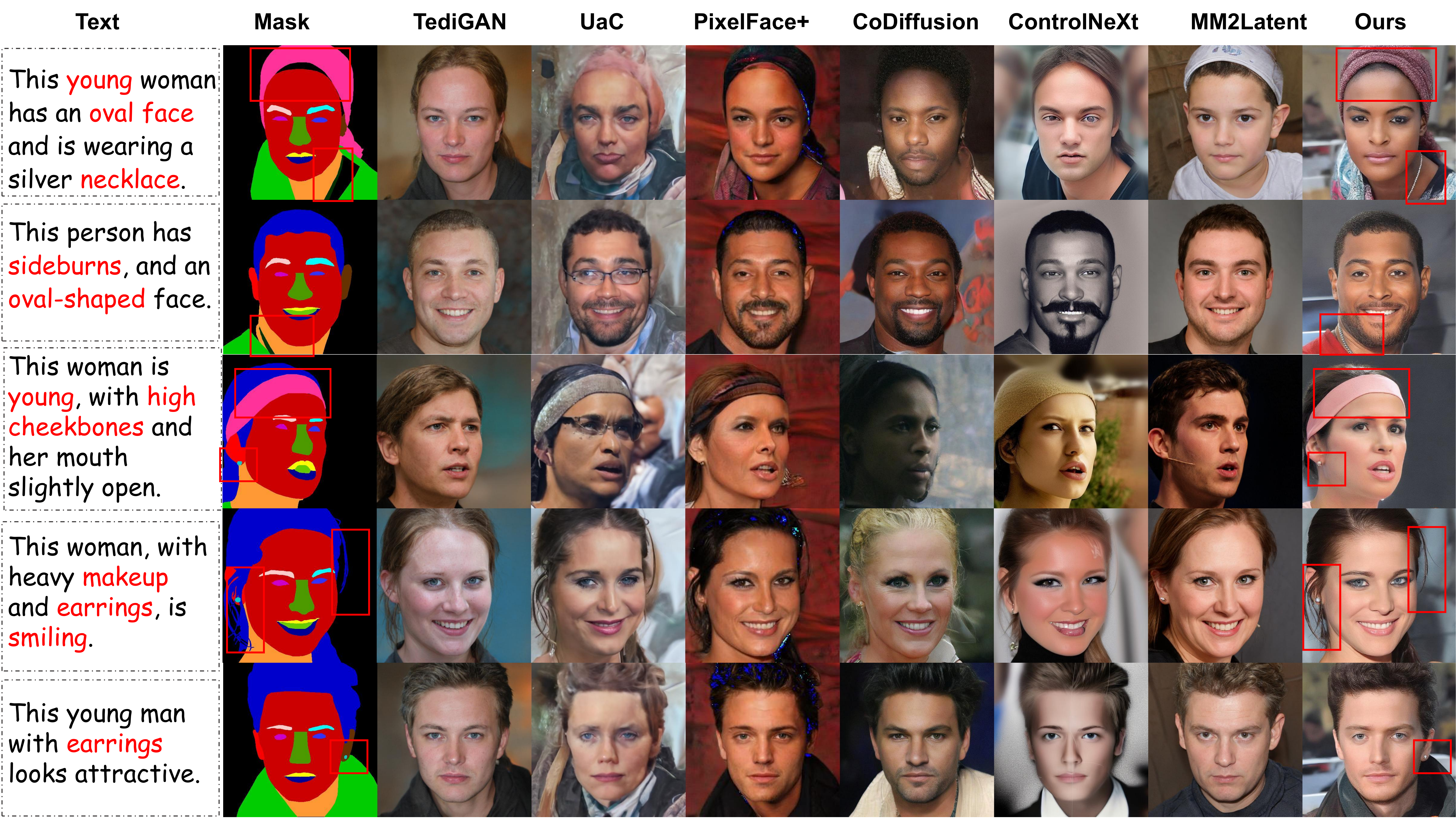}
    \caption{Qualitative comparison with state-of-the-art methods under joint text and mask guidance. EC\textsuperscript{2}Face generates images with better multimodal consistency, especially for fine-grained attributes such as hats, earrings, and necklaces.} 
    \label{fig:mmgen-vis}
\end{figure*}

\subsection{Experimental Setup}

\textbf{Datasets.}
We evaluate EC\textsuperscript{2}Face on MM-CelebA-HQ \cite{lee2020maskgan}, a widely used benchmark for multimodal face synthesis. The dataset contains 30,000 high-resolution RGB face images, each paired with 10 textual descriptions and 19-category semantic mask annotations. These annotations cover both facial regions, such as skin and facial components, and accessory attributes, such as earrings and necklaces. Following a 9:1 split, we divide the dataset into training and test sets for model learning and evaluation, respectively. To further assess cross-dataset generalization, we additionally construct a test set named MM-FFHQ by applying a pre-trained face parser \cite{zheng2022general} to FFHQ-Text \cite{zhou2021generative} to obtain semantic masks.

\textbf{Evaluation Metrics.}
We adopt widely used metrics to evaluate multimodal alignment and image quality. For text-image semantic consistency, we report the CLIP score \cite{radford2021learning}. For mask-image spatial consistency, we compute the mean accuracy (mAcc) over semantic regions. To assess image quality, we use LPIPS \cite{zhang2018unreasonable} to measure perceptual similarity between generated and real images, and NIQE \cite{mittal2012making} to evaluate the naturalness of synthesized images in a no-reference manner.

\textbf{Implementation Details.}
All experiments are conducted on a server with eight NVIDIA A100 GPUs (80GB each). We train the model using AdamW with an initial learning rate of $1\times10^{-4}$. Each GPU processes a mini-batch of four samples, resulting in an effective batch size of 32. The model is trained for 5,000 optimization steps with gradient accumulation enabled for stable convergence. Unless otherwise specified, the loss weights are set to $\alpha_1=\alpha_2=0.5$. During inference, we employ the Flow-Matching Euler Discrete scheduler \cite{esser2024scaling} with 28 sampling steps. Meanwhile, to ensure the reproducibility of results, we fix the random seed at 42.

\begin{table*}[ht]
\centering
\normalsize
\renewcommand{\arraystretch}{1.0}
\setlength{\tabcolsep}{6pt} 
\caption{Quantitative comparison with state-of-the-art methods on MM-CelebA-HQ. EC\textsuperscript{2}Face achieves the best performance in both text consistency and mask consistency, while also delivering competitive image quality.}
\label{tab:exp_mm-celeba}
\begin{tabular}{lcccccccc}
\toprule
\multirow{3}{*}[-0.5em]{Methods} 
& \multicolumn{5}{c}{Multimodal Conditional Consistency}  & \multicolumn{2}{c}{Visual Quality}  \\
\cmidrule(lr){2-6} \cmidrule(lr){7-8}
 & \multirow{2}{*}{\makecell[c]{CLIP-S (\%, $\uparrow$)}} 
 & \multicolumn{4}{c}{Mask mAcc (\%, $\uparrow$)}
& \multirow{2}{*}{\makecell[c]{LPIPS ($\downarrow$)}} & \multirow{2}{*}{\makecell[c]{NIQE ($\downarrow$)}} \\
\cmidrule(lr){3-6}
 &  & OAA & HFA & NFA & RFA   &  &   &    \\
\midrule
TediGAN~\cite{tedigan}   & 23.90 & 66.58 & 91.14  & 69.18  & 0.61  & \textbf{0.49}  &  5.19 \\
UaC~\cite{uniteandconquer}    & 25.52 & 65.37 & 79.54  & 68.27  & 18.19  & 0.57  & 6.47 \\
PixelFace+~\cite{pixelfaceplus}   & {26.16} & {82.11}    & 93.96     & 86.80     & 27.81  & 0.56  & 6.24  \\
CoDiffusion~\cite{collaborative} & 24.51 & 61.51 & 85.34  & 62.83  & 5.26   & 0.58  & 4.46 \\
ControlNeXt~\cite{peng2024controlnext}  & 25.88 & 80.93 & 91.67  & 84.94  & 33.40 & 0.54  & 4.54 \\
MM2Latent~\cite{meng2025mm2latent}  & 24.61 & 65.62 & 85.75  & 69.18  & 2.27  & {0.51}  & 3.74 \\
\midrule
\rowcolor{blue!5}
\textbf{EC\textsuperscript{2}Face (Ours)}  & \textbf{27.05}{\scriptsize(\textbf{\textcolor{red}{+0.89})}} & \textbf{88.64}{\scriptsize(\textbf{\textcolor{red}{+6.53})}} & \textbf{96.56}{\scriptsize(\textbf{\textcolor{red}{+1.6})}} & \textbf{90.18}{\scriptsize(\textbf{\textcolor{red}{+3.38})}}  & \textbf{62.78}{\scriptsize(\textbf{\textcolor{red}{\underline{+29.38}})}}   &  {0.51} & \textbf{3.55}  \\
\bottomrule
\end{tabular}
\end{table*}
\begin{table*}[ht]
\centering
\normalsize
\renewcommand{\arraystretch}{1.}
\setlength{\tabcolsep}{6pt} 
\caption{Quantitative results of EC\textsuperscript{2}Face and competing methods on MM-FFHQ. The proposed method consistently achieves superior performance, demonstrating strong cross-dataset robustness and generalization ability.}
\label{tab:exp_mm-ffhq}
\begin{tabular}{lcccccccc}
\toprule
\makecell[c]{\multirow{3}{*}[-0.5em]{Methods}} 
& \multicolumn{5}{c}{Multimodal Conditional Consistency}  & \multicolumn{2}{c}{Visual Quality}  \\
\cmidrule(lr){2-6} \cmidrule(lr){7-8}
 & \multicolumn{1}{c}{\multirow{2}{*}{CLIP-S (\%, $\uparrow$)}} 
 & \multicolumn{4}{c}{Mask mAcc(\%, $\uparrow$)}
& \multirow{2}{*}{LPIPS ($\downarrow$)} & \multirow{2}{*}{NIQE ($\downarrow$)} \\
\cmidrule(lr){3-6}
 &  & OAA & HFA & NFA & RFA   &  &   &    \\
\midrule
TediGAN~\cite{tedigan}   & 25.03   & 64.63   &  91.40  &  66.25   &  0.56  & 0.64  &  5.28  \\
UaC~\cite{uniteandconquer}    &  26.97  & 67.61   &  82.52   & 70.23    & 20.73    & 0.67  & 6.39   \\
PixelFace+~\cite{pixelfaceplus}   & 26.60  & 83.36   & 93.72   & 88.24   & 30.99   &  0.67   & 5.93   \\
CoDiffusion~\cite{collaborative} & 23.03   & 58.35   & 82.80   & 59.31    &  3.22   &  0.68   & 6.79   \\
ControlNeXt~\cite{peng2024controlnext}   & 27.80   &  71.63   & 86.93   & 74.72   &  20.95   &  0.62 & 3.76 \\
MM2Latent~\cite{meng2025mm2latent}  & 26.37   & 63.25  &  84.97   &  66.00  &   1.93  &  0.64   & 3.83   \\
\midrule
\rowcolor{blue!5}
\textbf{EC\textsuperscript{2}Face (Ours)}  & \textbf{28.40}{\scriptsize(\textbf{\textcolor{red}{+0.6})}} & \textbf{88.28}{\scriptsize(\textbf{\textcolor{red}{+4.52})}}  & \textbf{96.98}{\scriptsize(\textbf{\textcolor{red}{+3.26})}}  & \textbf{90.02}{\scriptsize(\textbf{\textcolor{red}{+1.78})}}  & \textbf{59.63}{\scriptsize(\textbf{\textcolor{red}{\underline{+28.64}})}}  & \textbf{0.51}  &  \textbf{3.67} \\
\bottomrule
\end{tabular}
\end{table*}
\subsection{Quantitative Comparison}
To provide a more fine-grained analysis of generation accuracy across attributes with different frequency characteristics, we evaluate mask-conditioned consistency not only in terms of the overall average accuracy (OAA), but also by separately reporting the mean accuracies of high-frequency attributes (HFA), normal-frequency attributes (NFA), and rare-frequency attributes (RFA). Specifically, attributes with both high average pixel proportion and high occurrence frequency are classified as HFA, while those with low values on both statistics are categorized as RFA; the remaining attributes are grouped into NFA. The detailed attribute partition is as follows.
\begin{itemize}
    \item \textbf{HFA}: 
        \texttt{background}, \texttt{hair}, \texttt{skin}, \texttt{neck}.
    \item \textbf{NFA}: 
        \texttt{cloth}, \texttt{nose}, \texttt{l\_lip}, \texttt{u\_lip}, \texttt{l\_ear}, \texttt{r\_ear}, 
        \texttt{l\_brow}, \texttt{r\_brow}, \texttt{mouth}, \texttt{l\_eye}, \texttt{r\_eye}, \texttt{hat}, \texttt{glasses}.
    \item \textbf{RFA}: 
        \texttt{earring}, \texttt{necklace}.
\end{itemize}
Table~\ref{tab:exp_mm-celeba} presents the quantitative comparison between EC\textsuperscript{2}Face and state-of-the-art methods on the MM-CelebA-HQ benchmark. The results show that EC\textsuperscript{2}Face achieves the best performance in both text consistency and mask consistency. In particular, for rare attributes, our method achieves a significant improvement, with its accuracy rate increasing by 29.38\% compared to that of the second-best performer. Meanwhile, it also achieves state-of-the-art performance in generated image quality. Table~\ref{tab:exp_mm-ffhq} further reports the quantitative results on the MM-FFHQ dataset. EC\textsuperscript{2}Face consistently outperforms competing methods, demonstrating strong cross-dataset robustness and generalization ability.

\subsection{Qualitative Comparison}
Figure~\ref{fig:mmgen-vis} presents visual comparisons between EC\textsuperscript{2}Face and prior state-of-the-art methods under joint text and mask guidance. The results show that EC\textsuperscript{2}Face achieves superior multimodal consistency by generating images that better satisfy both textual semantics and spatial mask constraints. Its advantage is particularly clear for fine-grained attributes, such as hats (row 1 and 3), earrings (rows 3--5), and necklaces (rows 1--2), where competing methods often exhibit missing details or mismatched semantics. By contrast, EC\textsuperscript{2}Face produces more accurate and visually coherent results, demonstrating stronger controllability under multimodal conditions. Additional qualitative results are provided in the Appendix.
\begin{figure*}[t]
    \centering
    \includegraphics[width=\textwidth]{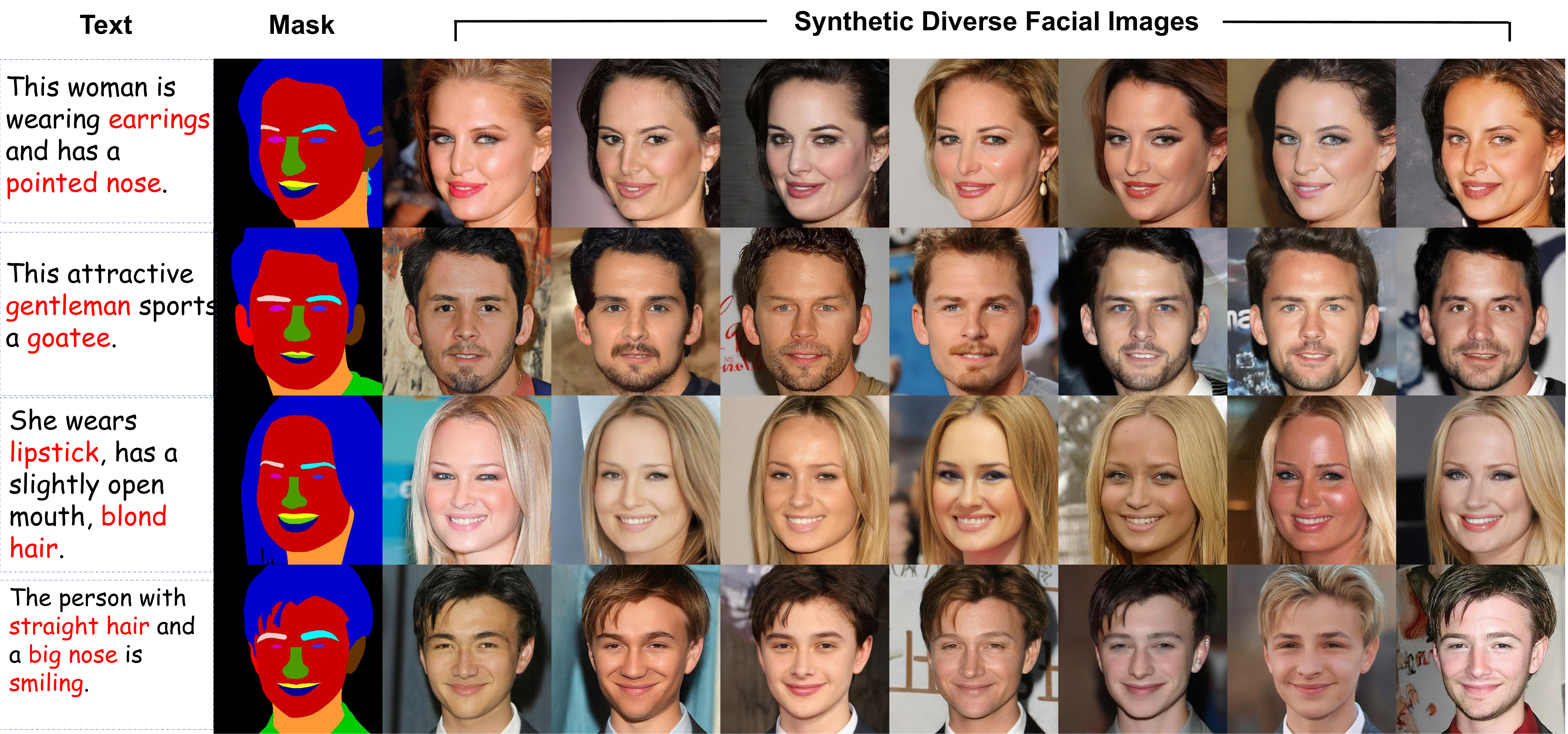}
    \caption{Generative diversity of EC\textsuperscript{2}Face under fixed text and mask conditions. By varying the random seed, the model produces diverse face images while preserving semantic consistency with the given conditions.}
    \label{fig:diversity}
\end{figure*}
\subsection{Generative Diversity}
We further investigate the generative diversity of EC\textsuperscript{2}Face under multimodal conditions. Specifically, with fixed text and mask inputs, we generate multiple face images by varying the random seed, and representative examples are shown in Figure~\ref{fig:diversity}. The results show that EC\textsuperscript{2}Face preserves accurate semantic consistency with the given conditions while maintaining considerable diversity in unconstrained attributes, such as identity, hair color, skin tone, and earring style. These findings indicate that the proposed method is capable of achieving both strong controllability and high diversity, which is highly beneficial for real-world applications such as data augmentation and interactive content creation.
\begin{figure}[htbp]
    \centering
    \begin{subfigure}[t]{\linewidth}
        \centering
        \includegraphics[width=0.95\linewidth]{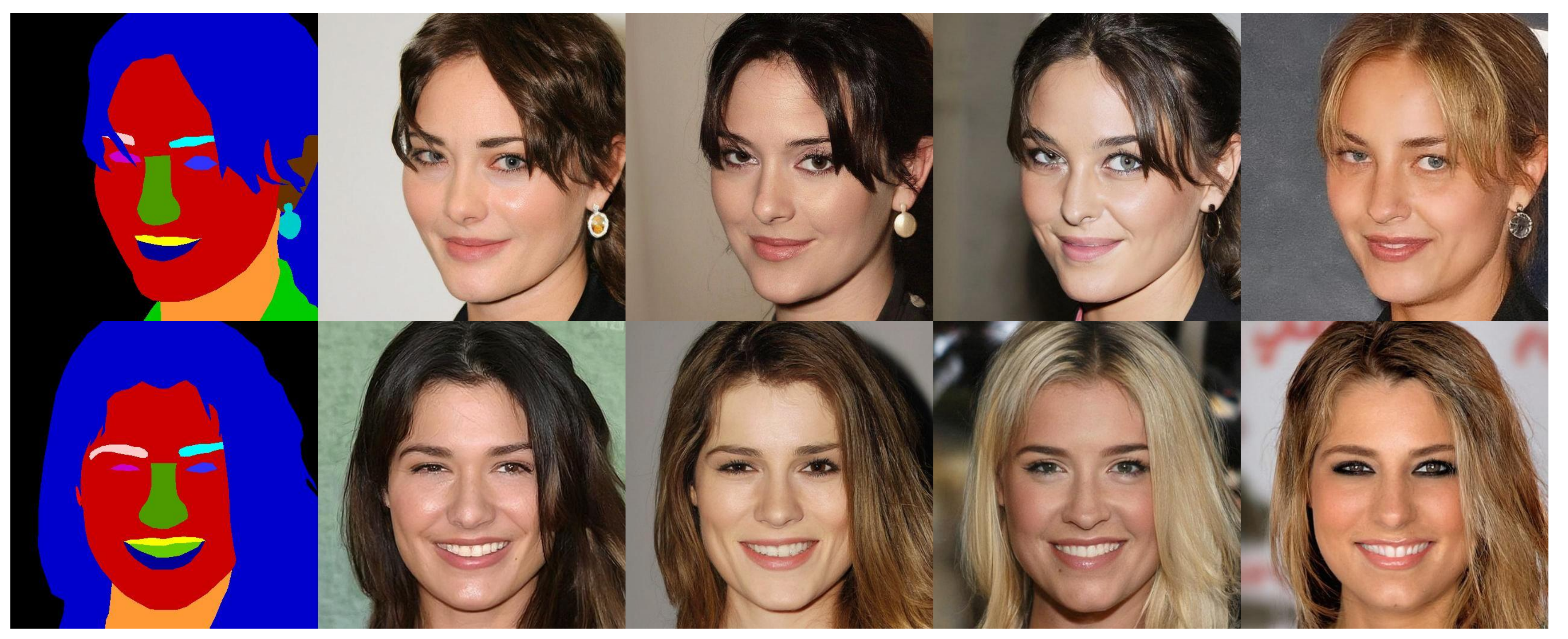} 
        \caption{Mask-driven only facial synthesis.}
    \end{subfigure}
    \par\vspace{0.1cm}
    \begin{subfigure}[t]{\linewidth}
        \centering
        \includegraphics[width=0.95\linewidth]{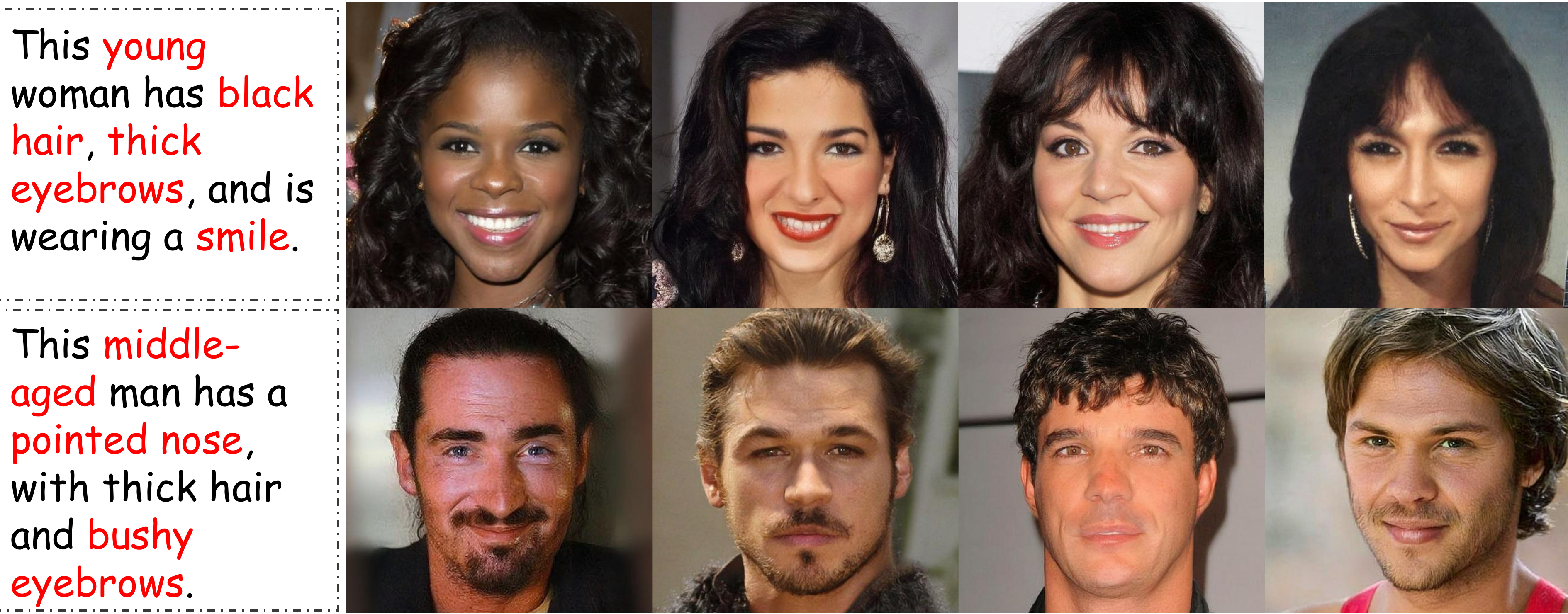} 
        \caption{Text-driven only facial synthesis.}
    \end{subfigure}
    \caption{EC\textsuperscript{2}Face showcases remarkable robustness in uni-modal (mask- or text-driven) facial synthesis.}
    \label{fig:unimodal2face}
\end{figure}
\subsection{Unimodal Conditional Generation}
EC\textsuperscript{2}Face not only excels in multimodal-collaborative facial synthesis but also demonstrates exceptional adaptability and robustness when faced with incomplete input conditions, such as providing only mask or text. As depicted in Fig. \ref{fig:unimodal2face}, when only a mask is input, the model generates facial images that not only maintain a high degree of spatial alignment with the mask but also exhibit a rich diversity in attributes, including identity and hair color. Conversely, when only text is input, the model ensures the accuracy of specific attributes while producing facial results with diverse spatial layouts, such as varied poses and expressions. This characteristic enables the model to adapt to a wider range of complex and ever-changing practical application requirements.
\subsection{Ablation Studies}
\begin{figure}[!t]
        \centering    
      \includegraphics[width=0.9\linewidth]{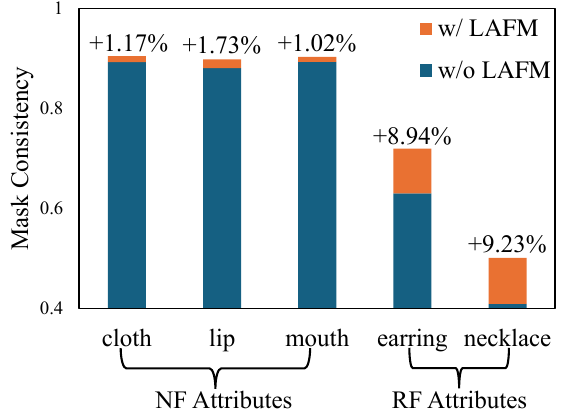}
        \caption{The use of LAFM significantly enhances the generation accuracy of rare attributes.}
        \label{fig:long_tail}
\end{figure}

\textbf{Effectiveness of LAFM.}
We first evaluate the effectiveness of Long-Tail Adaptive Flow Matching. As shown in Table~\ref{tab:LAL_effect}, LAFM significantly improves mask alignment for rare attributes, boosting the average accuracy from 52.04\% to 61.12\%. In contrast, the improvement on high-frequency attributes is relatively small, while the average accuracy on normal-frequency attributes increases by 0.79\%. Figure~\ref{fig:long_tail} clearly demonstrates, through the presentation of fine-grained accuracy comparison results, the notable advantages of LAFM in handling rare attributes.
\begin{table}[t]
    \centering
    \caption{Ablation study on the effectiveness of LAFM. Introducing LAFM significantly improves mask alignment performance, especially for rare attributes.}
    \label{tab:LAL_effect}
    \setlength{\tabcolsep}{7pt}
    \normalsize
    \renewcommand{\arraystretch}{1.}
    \begin{tabular}{ccccc}
    \toprule
    \multirow{2}{*}{\makecell{LAFM}} & \multicolumn{4}{c}{Mask mAcc(\%, $\uparrow$) } \\
    \cmidrule(lr){2-5}
     & OAA & HFA & NFA & RFA \\
    \midrule
    \XSolidBrush  & 86.14 & 95.13 & 88.62 & 52.04 \\
    \rowcolor{blue!5}
    \Checkmark   & \textbf{87.64} 
         & \textbf{95.18} 
         & \textbf{89.41} 
         & \textbf{61.12} {\scriptsize(\textbf{\textcolor{red}{+9.08})}} \\
    \bottomrule
    \end{tabular}
\end{table}

\textbf{Effectiveness of ECCG.}
Table~\ref{tab:ECCG_effect} presents an ablation study on Explicit Conditional Consistency Guidance. We consider four settings: removing both the text consistency loss $\mathcal{L}_{\text{text}}$ and the mask consistency loss $\mathcal{L}_{\text{mask}}$, using only $\mathcal{L}_{\text{mask}}$, using only $\mathcal{L}_{\text{text}}$, and using both simultaneously. Applying either loss alone improves semantic alignment in its corresponding modality, while combining both leads to consistent gains across all multimodal consistency metrics. These results demonstrate the effectiveness of ECCG and confirm that explicit consistency supervision in pixel space is beneficial for improving both modality-specific alignment and overall multimodal consistency.

\begin{table}[t]
    \centering
    \caption{Ablation study on ECCG. Both $\mathcal{L}_{\text{text}}$ and $\mathcal{L}_{\text{mask}}$ improve semantic alignment in their respective modalities, and their joint use simultaneously improves overall multimodal consistency.}
    \label{tab:ECCG_effect}
    \setlength{\tabcolsep}{3pt} 
    \normalsize
    \renewcommand{\arraystretch}{1.}
    \begin{tabular}{ccccccc} 
    \toprule
    \multicolumn{2}{c}{ECCG} & \multirow{2}{*}{CLIP-S (\%, $\uparrow$)} & \multicolumn{4}{c}{Mask mAcc(\%, $\uparrow$) } \\ 
    \cmidrule(lr){1-2} \cmidrule(lr){4-7}
    $\mathcal{L}_{\text{text}}$ & $\mathcal{L}_{\text{mask}}$  &  & OAA & HFA & NFA  & RFA \\ 
    \midrule
     \XSolidBrush  & \XSolidBrush  & 26.12  & 87.64 & 95.18 & 89.41 & 61.12 \\ 
     \XSolidBrush  & \Checkmark   & 26.26  & 88.59  & 96.50  & 90.14  &  62.68 \\ 
     \Checkmark   & \XSolidBrush  &  {27.22}  &  87.45 &  95.36 & 89.16  & 60.60  \\ 
    \rowcolor{blue!5}
    \Checkmark &  \Checkmark  & 27.05  & {88.64} & {96.56} & {90.18} & {62.78} \\ 
    \bottomrule
    \end{tabular}
\end{table}

\begin{table}[t]
    \centering
    \caption{Ablation study on temporal dynamic modulation. We compare the full model with a variant using a fixed modulation strength, i.e., $\beta(t)\equiv 1$.}
    \label{tab:validation_of_beta_t}
    \setlength{\tabcolsep}{3pt}
    \normalsize
    \renewcommand{\arraystretch}{1.}
    \begin{tabular}{ccccc}
    \toprule
    $\beta(t)$ & CLIP-S(\%,$\uparrow$) & mAcc(\%, $\uparrow$) & LPIPS($\downarrow$) & NIQE($\downarrow$) \\
    \midrule
    $\equiv 1$  & \textbf{28.45} & 88.04 & 0.57 & 3.88 \\
    \rowcolor{blue!5}
    Dynamic   & 27.05 
         & \textbf{88.64} 
         & \textbf{0.51}
         & \textbf{3.55}  \\
    \bottomrule
    \end{tabular}
\end{table}
\begin{table}[ht]
    \centering
    \caption{Influence of the hyperparameter $\alpha_1$.}
    \label{tab:validation_of_alpha_1}
    \setlength{\tabcolsep}{6pt}
    \normalsize
    \begin{tabular}{ccc}
    \toprule
    $\alpha_1$ & CLIP-S(\%, $\uparrow$) & LPIPS($\downarrow$) \\
    \midrule
   0. & 26.26  & 0.504 \\
   \rowcolor{blue!5}
   0.5 & 27.05 & 0.510 \\
   1.0 & 27.99 & 0.526 \\
   2.0 & 28.81 & 0.545 \\
    \bottomrule
    \end{tabular}
\end{table}
\begin{table}[ht]
    \centering
    \caption{Influence of the hyperparameter $\alpha_2$.}
    \label{tab:validation_of_alpha_2}
    \setlength{\tabcolsep}{6pt}
    \normalsize
    \begin{tabular}{ccc}
    \toprule
    $\alpha_2$ & mAcc(\%, $\uparrow$) & LPIPS($\downarrow$) \\
    \midrule
   0. & 87.45 & 0.515 \\
   \rowcolor{blue!5}
   0.5 & 88.64 & 0.510 \\
   1.0 & 90.15 & 0.519 \\
   2.0 & 91.03 &  0.535\\
    \bottomrule
    \end{tabular}
\end{table}

\begin{figure}[t] 
    \centering
    \includegraphics[width=0.95\linewidth]{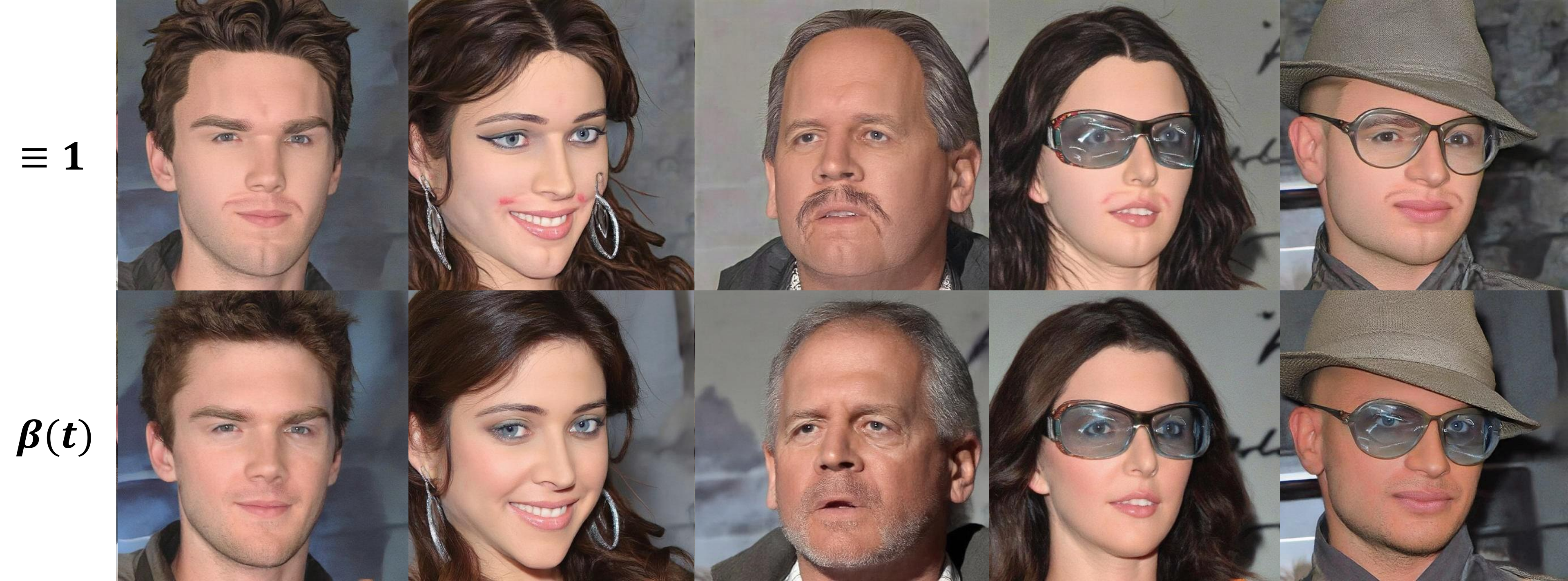}
    \caption{Visual comparison of generated faces with and without temporal dynamic modulation. Dynamic modulation of $\beta(t)$ produces more natural and visually realistic results under the same random seed.}
    \label{fig:wo_beta_t}
\end{figure}

\textbf{Effectiveness of Temporal Dynamic Modulation.}
To validate the temporal dynamic modulation function $\beta(t)$, we compare the full model with a variant that disables modulation by setting $\beta(t)\equiv 1$. The quantitative results are reported in Table~\ref{tab:validation_of_beta_t}. Without temporal modulation, the model still achieves strong multimodal consistency, and the text consistency score even slightly improves. However, LPIPS and NIQE deteriorate noticeably, indicating that imposing explicit consistency constraints with a fixed strength can interfere with the underlying generative process when reverse estimation is unreliable. Figure~\ref{fig:wo_beta_t} further shows that the model with dynamic modulation generates more natural and visually plausible faces. These results suggest that $\beta(t)$ effectively balances semantic alignment and image quality.

\textbf{Effects of Hyperparameters $\alpha_1$ and $\alpha_2$.}
We further study the influence of the loss weights $\alpha_1$ and $\alpha_2$, with results summarized in Tables~\ref{tab:validation_of_alpha_1} and~\ref{tab:validation_of_alpha_2}. Increasing either $\alpha_1$ or $\alpha_2$ consistently improves the semantic alignment metric of the corresponding modality, further confirming the benefit of explicit pixel-space consistency supervision. However, overly large values lead to noticeable degradation in perceptual image quality, as reflected by LPIPS. This suggests that excessively strong consistency constraints may disrupt the intrinsic generative capability of the backbone model. In practice, we set both $\alpha_1$ and $\alpha_2$ to 0.5, which provides a favorable trade-off between conditional consistency and generation quality.

\section{Conclusion}
\label{sec:conclusion}
This paper presented EC\textsuperscript{2}Face, a multimodal face synthesis framework for improving semantic alignment and rare-attribute modeling in controllable diffusion-based generation. By combining ECCG with LAFM, the proposed method strengthens semantic consistency with text and mask conditions while alleviating optimization bias caused by long-tailed semantic distributions. The temporal dynamic modulation mechanism further improves the robustness of pixel-space supervision across diffusion timesteps. Extensive experiments demonstrated that EC\textsuperscript{2}Face consistently outperforms strong baselines in generation quality, multimodal semantic alignment, and rare-attribute synthesis, achieving a 29.38\% improvement in mask accuracy on rare attributes. Since all proposed components are used only during training, EC\textsuperscript{2}Face introduces no additional computational overhead during inference, making it an effective and practical solution for high-fidelity multimodal face synthesis.

\bibliographystyle{IEEEtran}
\bibliography{main}

@inproceedings{stylegan,
  title={A style-based generator architecture for generative adversarial networks},
  author={Karras, Tero and Laine, Samuli and Aila, Timo},
  booktitle={CVPR},
  pages={4401--4410},
  year={2019}
}

@inproceedings{stylegan2,
  title={Analyzing and improving the image quality of stylegan},
  author={Karras, Tero and Laine, Samuli and Aittala, Miika and Hellsten, Janne and Lehtinen, Jaakko and Aila, Timo},
  booktitle={CVPR},
  pages={8110--8119},
  year={2020}
}

@inproceedings{styleclip,
  title={Styleclip: Text-driven manipulation of stylegan imagery},
  author={Patashnik, Or and Wu, Zongze and Shechtman, Eli and Cohen-Or, Daniel and Lischinski, Dani},
  booktitle={ICCV},
  pages={2085--2094},
  year={2021}
}

@inproceedings{cfg,
  title={Classifier-Free Diffusion Guidance},
  author={Ho, Jonathan and Salimans, Tim},
  booktitle={NeurIPSW},
  pages={1--8},
  year={2022}
}

@inproceedings{controlnet,
  title={Adding conditional control to text-to-image diffusion models},
  author={Zhang, Lvmin and Rao, Anyi and Agrawala, Maneesh},
  booktitle={ICCV},
  pages={3836--3847},
  year={2023}
}

@inproceedings{gligen,
  title={Gligen: Open-set grounded text-to-image generation},
  author={Li, Yuheng and Liu, Haotian and Wu, Qingyang and Mu, Fangzhou and Yang, Jianwei and Gao, Jianfeng and Li, Chunyuan and Lee, Yong Jae},
  booktitle={CVPR},
  pages={22511--22521},
  year={2023}
}

@article{ddpm,
  title={Denoising diffusion probabilistic models},
  author={Ho, Jonathan and Jain, Ajay and Abbeel, Pieter},
  journal={NeurIPS},
  volume={33},
  pages={6840--6851},
  year={2020},
}

@inproceedings{ddim,
  title={Denoising Diffusion Implicit Models},
  author={Song, Jiaming and Meng, Chenlin and Ermon, Stefano},
  booktitle={ICLR},
  pages={1--20},
  year={2021},
}

@inproceedings{ldm,
  title={High-resolution image synthesis with latent diffusion models},
  author={Rombach, Robin and Blattmann, Andreas and Lorenz, Dominik and Esser, Patrick and Ommer, Bj{\"o}rn},
  booktitle={CVPR},
  pages={10684--10695},
  year={2022},
}

@inproceedings{tedigan,
  title={Tedigan: Text-guided diverse face image generation and manipulation},
  author={Xia, Weihao and Yang, Yujiu and Xue, Jing-Hao and Wu, Baoyuan},
  booktitle={CVPR},
  pages={2256--2265},
  year={2021}
}

@inproceedings{uniteandconquer,
  title={Unite and conquer: Plug \& play multi-modal synthesis using diffusion models},
  author={Nair, Nithin Gopalakrishnan and Bandara, Wele Gedara Chaminda and Patel, Vishal M},
  booktitle={CVPR},
  pages={6070--6079},
  year={2023}
}

@inproceedings{collaborative,
  title={Collaborative diffusion for multi-modal face generation and editing},
  author={Huang, Ziqi and Chan, Kelvin CK and Jiang, Yuming and Liu, Ziwei},
  booktitle={CVPR},
  pages={6080--6090},
  year={2023}
}

@inproceedings{pixelfaceplus,
author = {Du, Xiaoxiong and Peng, Jun and Zhou, Yiyi and Zhang, Jinlu and Chen, Siting and Jiang, Guannan and Sun, Xiaoshuai and Ji, Rongrong},
title = {PixelFace+: Towards Controllable Face Generation and Manipulation with Text Descriptions and Segmentation Masks},
year = {2023},
booktitle = {ACM MM},
pages = {4666–4677},
numpages = {12},
}

@inproceedings{ddgi,
  title={Diffusion-driven gan inversion for multi-modal face image generation},
  author={Kim, Jihyun and Oh, Changjae and Do, Hoseok and Kim, Soohyun and Sohn, Kwanghoon},
  booktitle={CVPR},
  pages={10403--10412},
  year={2024}
}

@inproceedings{meng2025mm2latent,
  author  = "Meng, Debin and Tzelepis, Christos and Patras, Ioannis and Tzimiropoulos, Georgios",
  year    = 2025,
  title   = "{MM2Latent: Text-to-facial image generation and editing in GANs with multimodal assistance}",
  booktitle = "ECCV",
  pages   = "88-106",
  organization = "Springer"
}

@inproceedings{clip2latent,
author    = {Justin N. M. Pinkney and Chuan Li},
title     = {clip2latent: Text driven sampling of a pre-trained StyleGAN using denoising diffusion and CLIP},
booktitle = {BMVC},
year      = {2022},
pages     = {1--12}
}

@inproceedings{gcdp,
  title={Learning to generate semantic layouts for higher text-image correspondence in text-to-image synthesis},
  author={Park, Minho and Yun, Jooyeol and Choi, Seunghwan and Choo, Jaegul},
  booktitle={ICCV},
  pages={7591--7600},
  year={2023}
}

@inproceedings{e3facenet,
author = {Zhang, Jinlu and Zhou, Yiyi and Zheng, Qiancheng and Du, Xiaoxiong and Luo, Gen and Peng, Jun and Sun, Xiaoshuai and Ji, Rongrong},
title = {Fast text-to-3D-aware face generation and manipulation via direct cross-modal mapping and geometric regularization},
year = {2024},
pages = {60605--60625},
booktitle = {ICML},
numpages = {21},
}

@inproceedings{inade,
  title={Diverse semantic image synthesis via probability distribution modeling},
  author={Tan, Zhentao and Chai, Menglei and Chen, Dongdong and Liao, Jing and Chu, Qi and Liu, Bin and Hua, Gang and Yu, Nenghai},
  booktitle={CVPR},
  pages={7962--7971},
  year={2021}
}

@article{e2style,
  title={E2Style: Improve the efficiency and effectiveness of StyleGAN inversion},
  author={Wei, Tianyi and Chen, Dongdong and Zhou, Wenbo and Liao, Jing and Zhang, Weiming and Yuan, Lu and Hua, Gang and Yu, Nenghai},
  journal={TIP},
  volume={31},
  pages={3267--3280},
  year={2022},
}

@inproceedings{semflow,
  title={Semflow: Binding semantic segmentation and image synthesis via rectified flow},
  author={Wang, Chaoyang and Li, Xiangtai and Qi, Lu and Ding, Henghui and Tong, Yunhai and Yang, Ming-Hsuan},
  booktitle={NeurIPS},
  volume={37},
  pages={138981--139001},
  year={2024}
}

@article{peng2024controlnext,
  author  = "Peng, Bohao and Wang, Jian and Zhang, Yuechen and Li, Wenbo and Yang, Ming-Chang and Jia, Jiaya",
  year    = 2024,
  title   = "{Controlnext: Powerful and efficient control for image and video generation}",
  journal = "arXiv preprint arXiv:2408.06070"
}

@inproceedings{chen2024dreamidentity,
  author  = "Chen, Zhuowei and Fang, Shancheng and Liu, Wei and He, Qian and Huang, Mengqi and Mao, Zhendong",
  year    = 2024,
  title   = "{DreamIdentity: Enhanced editability for efficient face-identity preserved image generation}",
  booktitle = "AAAI",
  volume  = 38,
  number  = 2,
  pages   = "1281-1289"
}

@article{melnik2024face,
  author  = "Melnik, Andrew and Miasayedzenkau, Maksim and Makaravets, Dzianis and Pirshtuk, Dzianis and Akbulut, Eren and Holzmann, Dennis and Renusch, Tarek and Reichert, Gustav and Ritter, Helge",
  year    = 2024,
  title   = "{Face generation and editing with stylegan: A survey}",
  journal = "IEEE TPAMI",
  volume  = 46,
  number  = 5,
  pages   = "3557-3576",
  publisher = "IEEE"
}

@inproceedings{wang2025facea,
  author  = "Wang, Jiayu and Yu, Yue and Chen, Jingjing and Dai, Qi and Jiang, Yu-Gang",
  year    = 2025,
  title   = "{FaceA-Net: Facial Attribute-Driven ID Preserving Image Generation Network}",
  booktitle = "AAAI",
  volume  = 39,
  number  = 7,
  pages   = "7736-7743"
}

@article{ning2023multi,
  author  = "Ning, Xin and Nan, Fangzhe and Xu, Shaohui and Yu, Lina and Zhang, Liping",
  year    = 2023,
  title   = "{Multi-view frontal face image generation: A survey}",
  journal = "Concurrency and Computation: Practice and Experience",
  volume  = 35,
  number  = 18,
  pages   = "e6147"
}

@article{sowmya2024generative,
  author  = "Sowmya, BJ and Meeradevi, Seems Shedole",
  year    = 2024,
  title   = "{Generative adversarial networks with attentional multimodal for human face synthesis}",
  journal = "Indonesian Journal of Electrical Engineering and Computer Science",
  volume  = 33,
  number  = 2,
  pages   = "1205-1215"
}

@conference{du2023pixelface+,
  author  = "Du, Xiaoxiong and Peng, Jun and Zhou, Yiyi and Zhang, Jinlu and Chen, Siting and Jiang, Guannan and Sun, Xiaoshuai and Ji, Rongrong",
  year    = 2023,
  title   = "{Pixelface+: Towards controllable face generation and manipulation with text descriptions and segmentation masks}",
  booktitle = "ACM MM",
  pages   = "4666-4677"
}

@inproceedings{po2024state,
  author  = "Po, Ryan and Yifan, Wang and Golyanik, Vladislav and Aberman, Kfir and Barron, Jonathan T and Bermano, Amit and Chan, Eric and Dekel, Tali and Holynski, Aleksander and Kanazawa, Angjoo and others",
  year    = 2024,
  title   = "{State of the art on diffusion models for visual computing}",
  booktitle = "Computer Graphics Forum",
  volume  = 43,
  number  = 2,
  pages   = "e15063",
  organization = "Wiley Online Library"
}

@article{he2025diffusion,
  title={Diffusion models in low-level vision: A survey},
  author={He, Chunming and Shen, Yuqi and Fang, Chengyu and Xiao, Fengyang and Tang, Longxiang and Zhang, Yulun and Zuo, Wangmeng and Guo, Zhenhua and Li, Xiu},
  journal={IEEE TPAMI},
  year={2025},
}

@article{chang2025design,
  title={On the design fundamentals of diffusion models: A survey},
  author={Chang, Ziyi and Koulieris, George A and Chang, Hyung Jin and Shum, Hubert PH},
  journal={PR},
  pages={111934},
  year={2025},
}

@conference{mou2024t2i,
  author  = "Mou, Chong and Wang, Xintao and Xie, Liangbin and Wu, Yanze and Zhang, Jian and Qi, Zhongang and Shan, Ying",
  year    = 2024,
  title   = "{{T2i-adapter}: Learning adapters to dig out more controllable ability for text-to-image diffusion models}",
  booktitle = "AAAI",
  pages   = "4296-4304"
}

@conference{rombach2022high,
  author  = "Rombach, Robin and Blattmann, Andreas and Lorenz, Dominik and Esser, Patrick and Ommer, Bj{\"o}rn",
  year    = 2022,
  title   = "{High-resolution image synthesis with latent diffusion models}",
  booktitle = "IEEE/CVF CVPR",
  pages   = "10684-10695"
}

@article{dao2023flow,
  title={Flow matching in latent space},
  author={Dao, Quan and Phung, Hao and Nguyen, Binh and Tran, Anh},
  journal={arXiv preprint arXiv:2307.08698},
  year={2023}
}

@inproceedings{peebles2023scalable,
  title={Scalable diffusion models with transformers},
  author={Peebles, William and Xie, Saining},
  booktitle={IEEE/CVF ICCV},
  pages={4195--4205},
  year={2023}
}

@misc{flux2024,
  author  = "Black-Forest-Labs",
  year    = 2024,
  title   = "{FLUX}",
  howpublished = "\url{https://github.com/black-forest-labs/flux}"
}

@inproceedings{zheng2022general,
  author  = "Zheng, Yinglin and Yang, Hao and Zhang, Ting and Bao, Jianmin and Chen, Dongdong and Huang, Yangyu and Yuan, Lu and Chen, Dong and Zeng, Ming and Wen, Fang",
  year    = 2022,
  title   = "{General facial representation learning in a visual-linguistic manner}",
  booktitle = "IEEE/CVF CVPR",
  pages   = "18697-18709"
}

@inproceedings{radford2021learning,
  author  = "Radford, Alec and Kim, Jong Wook and Hallacy, Chris and Ramesh, Aditya and Goh, Gabriel and Agarwal, Sandhini and Sastry, Girish and Askell, Amanda and Mishkin, Pamela and Clark, Jack and others",
  year    = 2021,
  title   = "{Learning transferable visual models from natural language supervision}",
  booktitle = "ICML",
  pages   = "8748-8763",
  organization = "PmLR"
}

@inproceedings{milletari2016v,
  title={V-net: Fully convolutional neural networks for volumetric medical image segmentation},
  author={Milletari, Fausto and Navab, Nassir and Ahmadi, Seyed-Ahmad},
  booktitle={2016 fourth international conference on 3D vision (3DV)},
  pages={565--571},
  year={2016},
  organization={Ieee}
}

@inproceedings{lee2020maskgan,
  author  = "Lee, Cheng-Han and Liu, Ziwei and Wu, Lingyun and Luo, Ping",
  year    = 2020,
  title   = "{Maskgan: Towards diverse and interactive facial image manipulation}",
  booktitle = "IEEE/CVF CVPR",
  pages   = "5549-5558"
}

@article{hu_lora:_2021,
  author  = "Hu, Edward J and Shen, Yelong and Wallis, Phillip and Allen-Zhu, Zeyuan and Li, Yuanzhi and Wang, Shean and Wang, Lu and Chen, Weizhu",
  year    = 2021,
  title   = "{Lora: Low-rank adaptation of large language models}",
  journal = "arXiv preprint arXiv:2106.09685",
}

@inproceedings{esser2024scaling,
  title={Scaling rectified flow transformers for high-resolution image synthesis},
  author={Esser, Patrick and Kulal, Sumith and Blattmann, Andreas and Entezari, Rahim and M{\"u}ller, Jonas and Saini, Harry and Levi, Yam and Lorenz, Dominik and Sauer, Axel and Boesel, Frederic and others},
  booktitle={ICML},
  year={2024}
}

@inproceedings{zhang2018unreasonable,
  author  = "Zhang, Richard and Isola, Phillip and Efros, Alexei A and Shechtman, Eli and Wang, Oliver",
  year    = 2018,
  title   = "{The unreasonable effectiveness of deep features as a perceptual metric}",
  booktitle = "IEEE CVPR",
  pages   = "586-595"
}

@article{mittal2012making,
  title={Making a “completely blind” image quality analyzer},
  author={Mittal, Anish and Soundararajan, Rajiv and Bovik, Alan C},
  journal={IEEE Signal processing letters},
  volume={20},
  number={3},
  pages={209--212},
  year={2012},
  publisher={IEEE}
}

@conference{wang2025lavin,
  author  = "Wang, Zhaoqing and Xia, Xiaobo and Chen, Runnan and Yu, Dongdong and Wang, Changhu and Gong, Mingming and Liu, Tongliang",
  year    = 2025,
  title   = "{Lavin-dit: Large vision diffusion transformer}",
  booktitle = "CVPR",
  pages   = "20060-20070"
}

@conference{zhou2021generative,
  author  = "Zhou, Yutong",
  year    = 2021,
  title   = "{Generative adversarial network for text-to-face synthesis and manipulation}",
  booktitle = "ACM MM",
  pages   = "2940-2944"
}

@inproceedings{diao2025ft2tf,
  title={Ft2tf: First-person statement text-to-talking face generation},
  author={Diao, Xingjian and Cheng, Ming and Barrios, Wayner and Jin, SouYoung},
  booktitle={IEEE/CVF WACV},
  pages={4821--4830},
  year={2025},
  organization={IEEE}
}

@article{song2025attridiffuser,
  title={AttriDiffuser: Adversarially enhanced diffusion model for text-to-facial attribute image synthesis},
  author={Song, Wenfeng and Ye, Zhongyong and Sun, Meng and Hou, Xia and Li, Shuai and Hao, Aimin},
  journal={Pattern Recognition},
  volume={163},
  pages={111447},
  year={2025},
  publisher={Elsevier}
}

@inproceedings{wu2023high,
  title={High-fidelity 3d face generation from natural language descriptions},
  author={Wu, Menghua and Zhu, Hao and Huang, Linjia and Zhuang, Yiyu and Lu, Yuanxun and Cao, Xun},
  booktitle={Proceedings of the IEEE/CVF Conference on Computer Vision and Pattern Recognition},
  pages={4521--4530},
  year={2023}
}

@article{liu2023gan,
  title={Gan-based facial attribute manipulation},
  author={Liu, Yunfan and Li, Qi and Deng, Qiyao and Sun, Zhenan and Yang, Ming-Hsuan},
  journal={IEEE transactions on pattern analysis and machine intelligence},
  volume={45},
  number={12},
  pages={14590--14610},
  year={2023},
  publisher={IEEE}
}

@inproceedings{tan2025ominicontrol,
  title={Ominicontrol: Minimal and universal control for diffusion transformer},
  author={Tan, Zhenxiong and Liu, Songhua and Yang, Xingyi and Xue, Qiaochu and Wang, Xinchao},
  booktitle={Proceedings of the IEEE/CVF International Conference on Computer Vision},
  pages={14940--14950},
  year={2025}
}

@inproceedings{cao2026multivariate,
  title={Multivariate diffusion transformer with decoupled attention for high-fidelity mask-text collaborative facial generation},
  author={Cao, Yushe and Shi, Dianxi and Fu, Xing and Zou, Xuechao and Peng, Haikuo and Li, Xueqi and Yu, Chun and Xing, Junliang},
  booktitle={Proceedings of the AAAI Conference on Artificial Intelligence},
  volume={40},
  number={4},
  pages={2670--2679},
  year={2026}
}

@inproceedings{cao2024mefusion,
  title={MEFusion: Unsupervised Mutual Enhancement for Multimodal Image Fusion},
  author={Cao, Yushe and Jiao, Siwen and Sun, Penghao and Peng, Baoyun and Shi, Dianxi and Shi, Yuanchun},
  booktitle={ECAI 2024: 27th European Conference on Artificial Intelligence, 19--24 October 2024, Santiago de Compostela, Spain--Including 13th Conference on Prestigious Applications of Intelligent Systems (PAIS 2024)},
  pages={553--560},
  year={2024},
  organization={SAGE Publications Pvt. Ltd 1 Oliver's Yard, 55 City Road, London, EC1Y 1SP}
}

@inproceedings{cao2026dual,
  title={Dual-Pathway Diffusion for Hand Correction in Synthetic Portraits: Global Context Aware and Local Structure Refinement},
  author={Cao, Yushe and Jing, Luoxi and Wang, Yuanze and Shi, Dianxi and Yu, Chun and Xing, Junliang},
  booktitle={Proceedings of the 2026 International Conference on Multimedia Retrieval},
  pages={1899--1907},
  year={2026}
}

\begin{figure*}[t]
    \centering    \includegraphics[width=\linewidth]{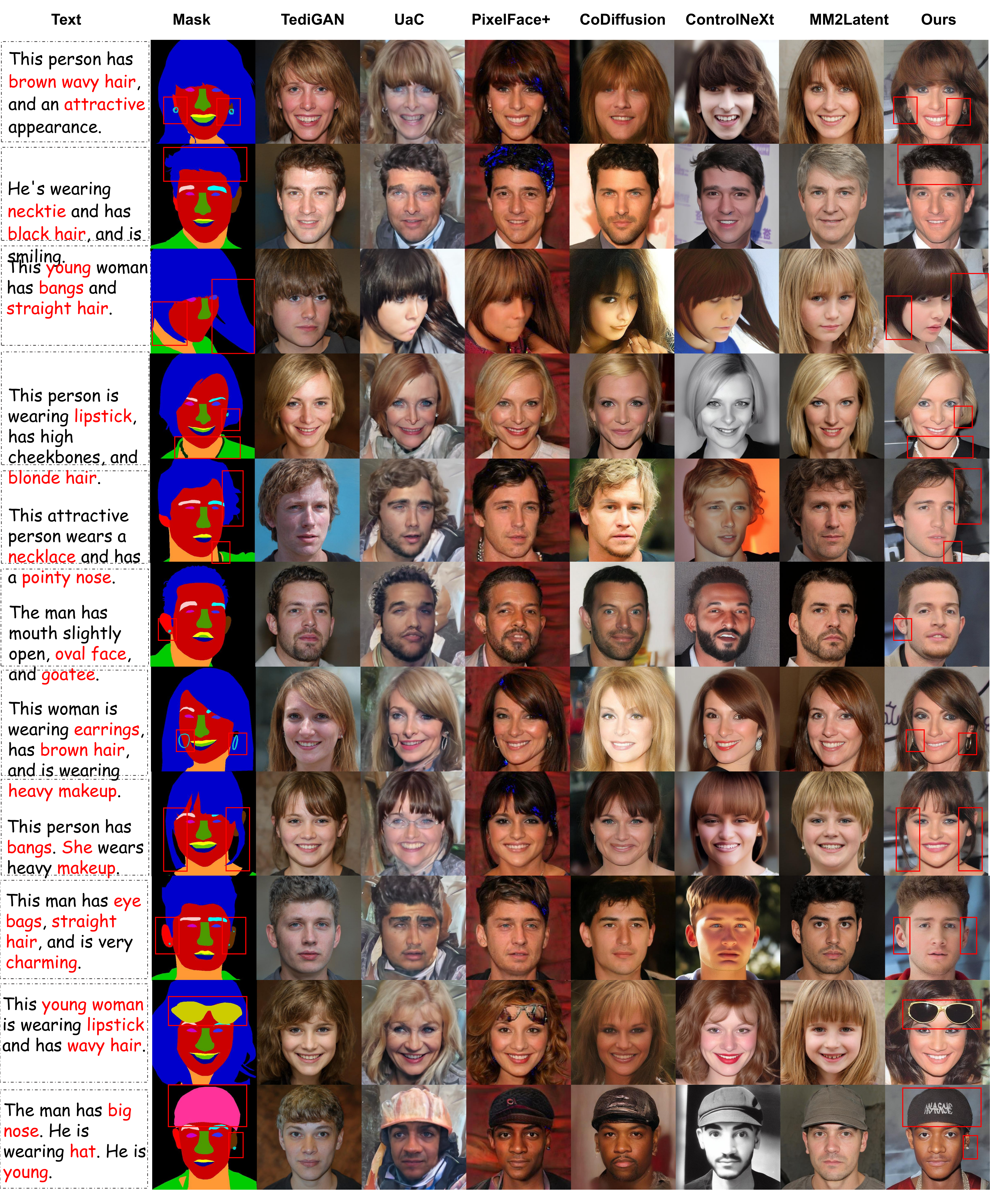}
    \caption{More visual comparisons with state-of-the-art methods: The facial images synthesized by EC\textsuperscript{2}Face significantly outperform competing methods in terms of multimodal consistency. Please zoom in for a better view.} 
    \label{fig:mmgen-vis}
\end{figure*}

\begin{figure*}[t]
    \centering    \includegraphics[width=\linewidth]{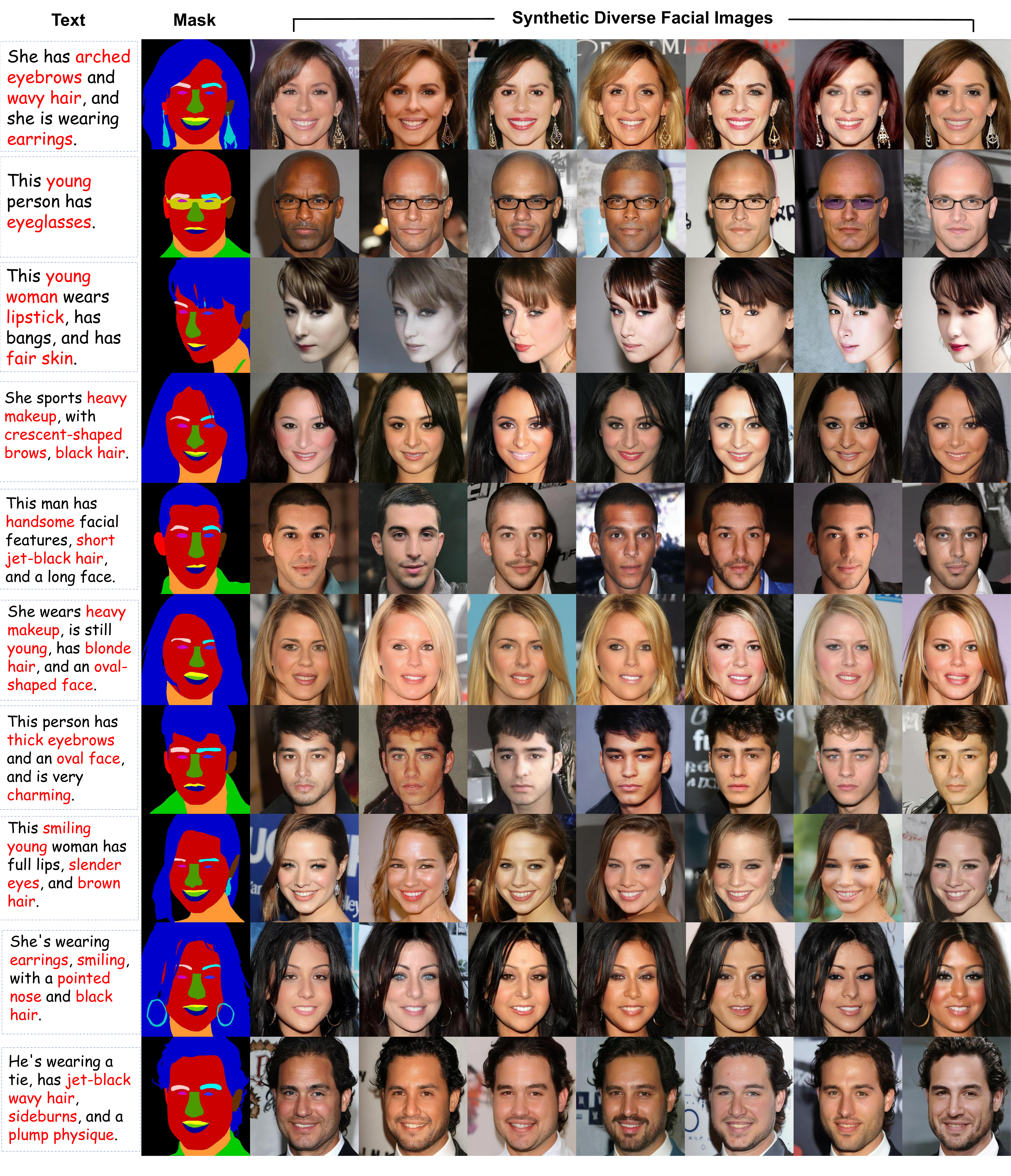}
    \caption{More examples of generated diversity: While maintaining consistency with the given conditions (mask and text), EC\textsuperscript{2}Face demonstrates high diversity in non-specified attributes such as identity, skin tone, and so on.} 
    \label{fig:mmgen-vis}
\end{figure*}

\begin{figure*}[t]
    \centering    \includegraphics[width=\linewidth]{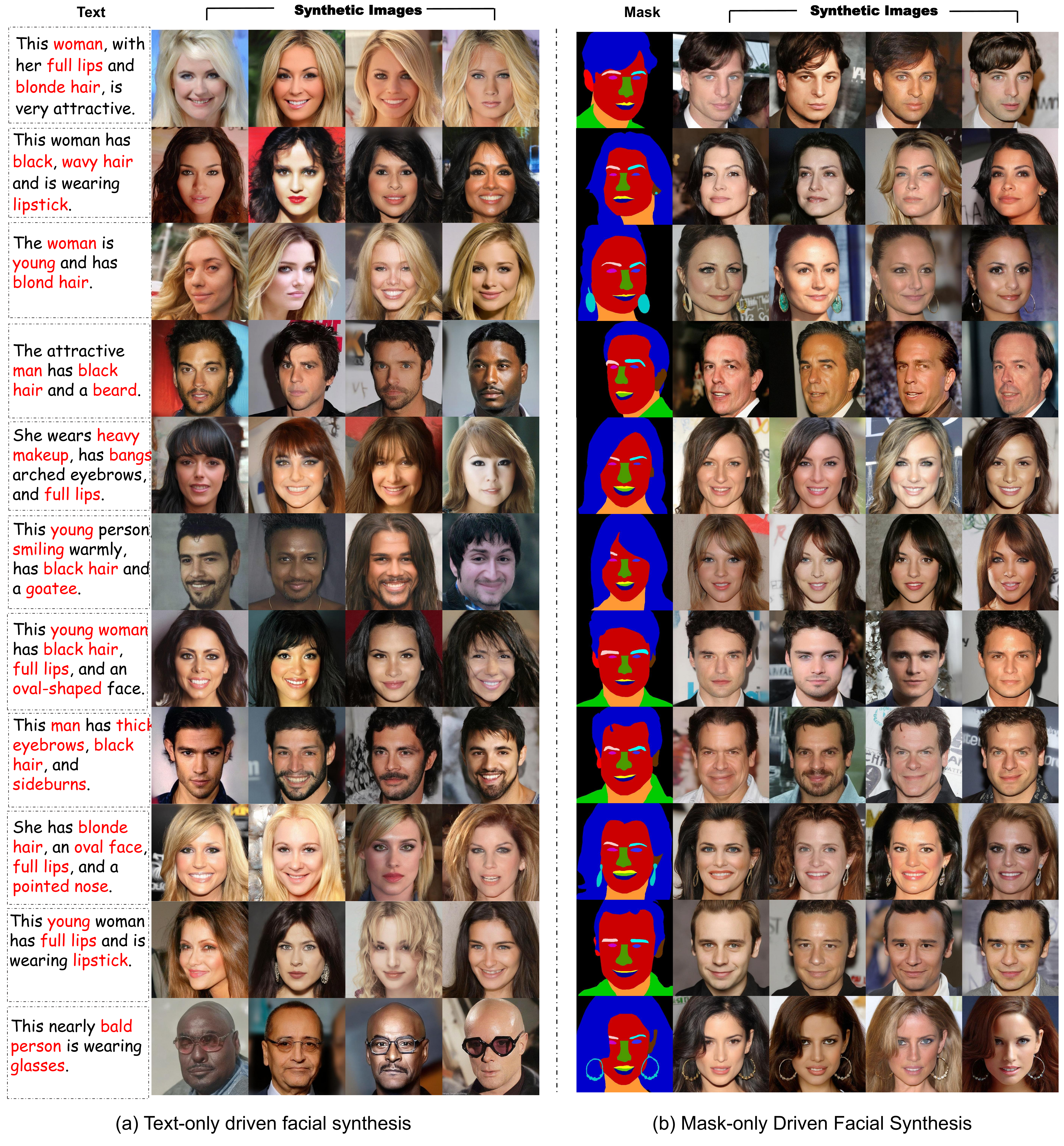}
    \caption{More examples of unimodal-driven facial synthesis: When confronted with incomplete conditions, such as being provided with only a mask or only text, EC\textsuperscript{2}Face is still capable of generating visually plausible facial images while exhibiting good consistency and diversity.} 
    \label{fig:mmgen-vis}
\end{figure*}
\vfill

\end{document}